\documentclass[%
 reprint,
 amsmath,amssymb,
 aps,
]{revtex4-1}
\usepackage{mathtools}
\usepackage{multirow}
\usepackage[export]{adjustbox}

\usepackage{graphicx}
\usepackage{dcolumn}
\usepackage{bm}
\usepackage{hyperref}
\usepackage[english]{babel}

\usepackage[normalem]{ulem}
\usepackage{comment}
\usepackage[draft]{todonotes}   
\usepackage{xcolor}
\usepackage{subcaption} 
\usepackage{booktabs} 
\begin{document}

\preprint{APS/123-QED}

\title{Neural Integration of Continuous Dynamics}

\author{Margaret Trautner}\email{trautner@mit.edu}
\affiliation{Department of Mathematics}
\author{Sai Ravela}
 \email{ravela@mit.edu}
\affiliation{Department of Earth, Atmospheric, and Planetary Sciences\\ \\ Earth Signals and Systems Group, Massachusetts Institute of Technology, Cambridge, MA 02139, USA}



\date{\today}

\begin{abstract}
Neural dynamical systems are dynamical systems that are described at least in part by neural networks. The class of continuous-time neural dynamical systems must, however, be numerically integrated for simulation and learning.  Here, we present a compact neural circuit for two common numerical integrators: the explicit Runge-Kutta method and the semi-implicit/predictor-corrector Adams-Bashforth-Moulton method. Modeled as recurrent networks embedding a continuous neural differential equation, they achieve fully neural temporal output. Using the polynomial class of dynamical systems, we demonstrate equivalence of neural and numerical integration.  

\end{abstract}

\maketitle

\section{Introduction}

Neural dynamical systems are dynamical systems described at least in part by neural networks. Our interest in the subject emerges in the context of Systems Dynamics and Optimization~\cite{dddas} (SDO), which is central to many  applications such as storm prediction~\cite{ravela12},   climate-risk based decision support~\cite{ravela10}, or autonomous observatories~\cite{ravela13}. The SDO cycle conceptually involves a forward path dynamically parameterizing, reducing, calibrating, initializing and simulating numerical models, and quantifying their uncertainties. SDO further involves a return path for adaptive observation, inversion and estimation.  In this context, neural networks are attractive as rapidly executing surrogates, and as parametrizations of dynamics difficult numerically model, such as sub-grid-scale turbulence~\cite{karpatne19}. 

Neural networks were classically developed to model discrete processes~\cite{lstm}, but physical systems evolve continuously in time. Continuous-time Neural Networks (CtNNs) seek to advance neural modeling in both feed-forward or recurrent forms; the latter consist of neurons connected through a time delay and continuous-time activation \cite{CtNN_Chen}. 

CtNNs can approximate  the time trajectory of any $n$-dimensional dynamical system to arbitrary accuracy~\cite{CtRNN_Funahashi}. Interest in them as Neural ODEs~\cite{neuode} has grown, particularly to describe dynamics in the form $\dot{x} = f_{NN}(x_t,u_t,t;w_t)$, with inputs $x_t$, feed-forward/feedback components $u_t$, and parameters (weights, biases) $w_t$ being a function of time $t\in\mathbb{R}$. More common is to consider autonomous dynamics, for example $\dot{x} = \dot{x}_{NUM}+ \dot{x}_{NN}$,  superposing a non-neural part ($f_{NUM}$) and a neural CtNN part ($f_{NN}$). The neural part may be modeled with time-invariant parameters, i.e. $\dot{x}_{NN} =  f_{NN}(x_t,u_t;w)$. 

Variational estimation for continuous dynamical systems directly applies to training Neural-ODEs. The attendant Hamilton-Jacobi-Bellman (HJB) equations estimate CtNN and non-neural parameters alike~\cite{bryson}. Although the Machine Learning community cites the use of HJB as a critical discovery~\cite{neuode}, nevertheless, HJB applies  to continuous dynamics in general, and requires no special adaptation for Neural-ODEs. In practice, it is essential  to consider the discrete adjoint equations whence a fundamental connection becomes even more plainly visible. For  discrete networks, the adjoint equations exactly define back-propagation~\cite{mlclass,bryson}. They emerge as normal equations of a multi-stage (multi-layer network) two-point boundary value problem~\cite{bryson} for stage parameters (layer weights and biases) using terminal constraints (training data). Simply changing the definition of stages from ``layers" to ``discrete (time) steps," the same methodology yields forward-backward iterations for training discretized CtNNs.

Whether CtNNs are interesting in their own right or as models of continuous processes, discrete-time simulation is presently central for training and operating them. The standard advice~\cite{neuode} is to use neural computation to calculate derivatives, then resort to a classical numerical integration, interleaving the two. Couldn't a purely neural computation generate the discrete solution instead?

In this first paper of a three-part series, we show that a single recurrent {\em Discrete-time Neural Network} (DtNN) couples derivative calculation and numerical stepping to generate numerical sequences, which we define as {\em neural integration.} We construct DtNNs to implement the explicit fixed-step Runge-Kutta method~\cite{press07} of arbitrary order. The key benefit of our approach is compactness; unlike extant methods~\cite{RKNN, RKNN2019}, the order of approximation for integration does not change the size of our network. We further apply this methodology to semi-implicit methods~\cite{press07}. Specifically, we couple a two-step Adams-Bashforth predictor to a two-step Adams-Moulton corrector. Once initialized, a single DtNN in each case performs neural integration, sequentially issuing outputs.  

To demonstrate our approach, we define and employ {\em PolyNets}, the class of CtNNs for dynamical systems with polynomial nonlinearities. PolyNets are exact when polynomial coefficients of the governing equations are known. Exact PolyNets conveniently prove correctness of the proposed neural integration methods. We further demonstrate correspondence between neural and numerical integration  using numerical simulations of the chaotic Lorenz-63~\cite{Lorenz1963} model. Extensions of methodology to non-polynomial dynamics are also discussed.  

The remainder of this paper is organized as follows. Related work is discussed in  Section~\ref{sec:rw}. PolyNets are discussed briefly in section~\ref{sec:polynet}. Neural integration is developed in Section~\ref{sec:neuint}, followed by a discussion in Section~\ref{sec:disc} and Conclusions in Section~\ref{sec:concl}.

\section{Related Work}
\label{sec:rw}

Multiple efforts for implementing neural ordinary differential equations~\cite{neuode} are being developed, including by incorporating physical constraints during learning~\cite{RackauckasDiffEq}.  Neural-ODEs must be discretized for numerical computation; it is typically suggested that CtNNs produce the derivatives subsequently passed to a traditional numerical integrator before returning to the neural computation as input for the next time step~\cite{neuode}.  Our approach instead allows the CtNN to be discretized and executed  entirely as a DtNN for neural integration.

Previous works have constructed numerical integrators in the form of a neural circuit~\cite{RKNN,RKNN2019}. However, the size of these networks scales with the order of the Runge-Kutta method.  Our approach offers a fixed recurrent architecture invariant to order, which is possible through the use of multiplicative nodes, which are now fairly common, e.g. see~\cite{lstm}. 

We devise methods for explicit Runge-Kutta (RK) of fixed time-step.  We also demonstrate a semi-implicit predictor-corrector Adams-Bashforth-Moulton (ABM) method.  To the best of our knowledge, the fixed-size RK neural integrator and the ABM neural integrator are new.  

Our neural integration approach is easily demonstrated as correct. Each neural integration method described here is a PolyNet. If the embedded CtNN is certified to be ``error-free" then so is neural integration;  no algorithmic error beyond the approximation inherent in the numerical method is induced. However, CtNN certification is generally difficult because  they are typically trained to an error tolerance. Except when CtNNs are exact PolyNets, the DtNN in our construction is also an exact PolyNet. Thus, algorithmic correctness of neural integration follows. This is important because differences between neural integration and numerical integration can in general arise for many reasons including training error, numerical approximation, computational implementation and numerical precision. It would be difficult to partition these errors, particularly when modeling chaotic processes. 

CtNNs must in general be trained and even in the special case of PolyNets,  this is true when the polynomial coefficients are unknown. Solving the discrete adjoint equations corresponding to HJB for learning requires discrete forward and backward numerical integration, which neural integration can accurately and efficaciously perform.

We posit that a key advantage of neural integration is that specialized hardware can be exploited better, particularly for Monte Carlo simulations of dynamical systems. For example, CUDA benefits of parallel simulations, e.g. for Monte Carlo, have already been shown~\cite{Niemeyer2014}.
 
\section{Exact PolyNets}
\label{sec:polynet}

To demonstrate neural integration, we introduce PolyNets, a class of CtNNs that  match polynomial dynamics. Polynomial dynamics informs a broad class ODEs and PDEs (e.g. Navier Stokes). Although we do not consider PDEs in this work, the methods described here also informs their numerical  integration. In Algorithm 1 that follows, we show that neural computation of polynomial dynamics with known coefficients is (trivially) exact. A fuller discussion of PolyNet classes is presented in Section~\ref{sec:disc}.

 Let $x\in\mathbb{R}^N$ be an $N$-dimensional variable and a dynamical system $\dot{x} = f(x;c)\in\mathbb{R}^N$, where $f$ is a polynomial containing monomials of maximum degree $D$. There are ${N+D \choose D}$ possible monomial terms per output and there are $N$ outputs. The function $f$ can be of the form:
 \begin{equation}
 \label{eq:polydyn}
     \dot{x}_n = c_{n,0}+\sum_{i=1}^N c_{n,i}x_i+\sum_{d=2}^D g(x,n,d),
 \end{equation}
 where, $n=1\ldots N$, and $g$ is defined as 
 \begin{equation}
     g(x,n,d) \doteq \sum_{i_1=1}^{N}\sum_{i_2=1}^{i_1}\ldots \sum_{i_d=1}^{i_{d-1}} c_{n,i_1\ldots i_d} x_{i_1}\cdots x_{i_d}
 \end{equation}
 
 The equivalent PolyNet 
 
 \begin{equation}
     \dot{x}=f_{NN}(x;w)
 \end{equation} 
 
 is defined as a three-layer network with $N$ inputs $x_1\ldots x_N$, $N$ outputs $\dot{x}_1\ldots \dot{x}_N$, and one hidden layer $h$. $f_{NN}$ is constructed as follows (Algorithm 1):
 \begin{enumerate}
     \item Set each output node's activation to be linear. \item Set $c_{n,0}$ as the bias for the output node $\dot{x}_n$. 
     \item Connect input node $x_i$ to output node  $\dot{x}_n$ with weight $c_{n,i}>0$,  where $i=1\ldots N$ and $n=1\ldots N$.
     \item To map the input to the hidden layer, re-write the single term, 
 \begin{equation}
 h_{i_1\ldots i_d}  \doteq  \prod_{j=1}^N x_j^{d_j}=x_{i_1}\cdots x_{i_d}  
 \end{equation}
 where $d_j = \sum_{k=1}^d \delta_{i_k, j}\in\mathbb{Z}^*$ and $\delta$ is the Kronecker delta function. 
 \item The term $h_{i_1\ldots i_d}$ is a hidden product node with linear activation and connecting to the input layer. Product nodes are commonly used in deep networks~\cite{lstm}. There are exactly $d_j$ connections between node $x_j$ and $h_{i_1\ldots i_d}$, each of unit weight. Note logarithms and anti-logarithms among others are obvious ways to express powers. however,  we've avoided them to  preclude invoking complex number representations. 
 \item A hidden node  $h_{i_1\ldots i_d}$ connects to an output node $\dot{x}_n$ additively with weight $c_{n,i_1\ldots i_d}>0$, where $n=1\ldots N$ and  $d=2\ldots D$.
 \item  There are at most $L= {N+D \choose D}-(1+N)$ nodes in the hidden layer. Thus the PolyNet with parameters $c\equiv c_{n,0},c_{n,i},c_{n,i_1\ldots i_d}$, $i,n=1\ldots N$, $d=2\ldots D$ and at most $nL$ weights of value $1$ constitute the parameter vector $w$ of $f_{NN}$.  
 \end{enumerate}
Thus, by Algorithm 1, the dynamical system $\dot{x}=f(x;c)$, (Equation~\ref{eq:polydyn}) and the PolyNet neural network $\dot{x}=f_{NN}(x;w)$ are equal. 

\subsection{Example: Lorenz Dynamics}

\begin{figure}[htb!]
{\centering
\includegraphics[width=0.55\linewidth]{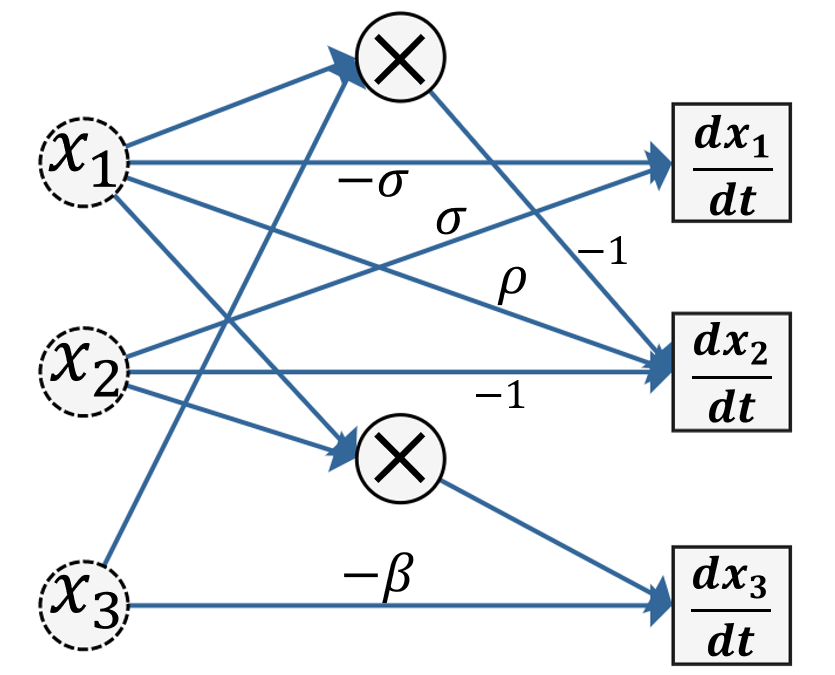}}
\caption{\it The Lorenz CtNN. Values adjacent to edges indicate weights, i.e. multiplication by that parameter along the edge. Edges without explicit labels are assumed to have weight of 1. Nodes with ``$\mathbf{\times}$" are multiplicative, nodes with ``\textbf{+}" are additive.}
\label{fig:Lorenz_Circuit}
\end{figure}

In Figure \ref{fig:Lorenz_Circuit}, a neural architecture derived directly from the governing equations of the Lorenz system computes the time derivatives given an input. The equations describing the system are
\begin{eqnarray*}
    \frac{dx_1}{dt} &=& \sigma(x_2-x_1)\\
    \frac{dx_2}{dt} &=& \rho x_1 - x_2 -x_1 x_3\\
    \frac{dx_3}{dt} &=& -\beta x_3 + x_1 x_2 .\\
\end{eqnarray*}

Interestingly, Lorenz system is a polynomial with $N=3$ variables and $D=2$ degree (maximum) but requires no more than two hidden nodes as implemented by the PolyNet. Note also that hidden nodes of degree $1$ are technically unnecessary, and we have simply replaced them with the necessary additive connections from input to output nodes.

\section{Neural Integration}
\label{sec:neuint}

In this section, we construct neural integrators, first for the Runge-Kutta (RK) method and second to Adams-Bashforth-Moulton (ABM). For the Runge-Kutta method, consider a neural implementation of the common $4^{\text{th}}$ order explicit Runge-Kutta scheme with fixed step:
\begin{eqnarray}
\label{eq:rk4a}
 k_1 &=& h f_{NN}(x_n)\\
 k_2 &=& h f_{NN}(x_n+\frac{1}{2} k_1) \\
 k_3 &=& h f_{NN}(x_n + \frac{1}{2} k_2) \\
 k_4  &=& h f_{NN}(x_n + k_3) \\
 \label{eq:rk4b} x_{n+1} &=& x_n + \frac{1}{6} (k_1 + 2\;k_2+ 2\;k_3 + k_4)
\end{eqnarray}

If $f_{NN}$ is an external function guaranteed to be accurate/correct, then the step-update is polynomial in $x_n$ and $k_i$, $i=1\ldots 4$ and it is an exact PolyNet. A simple design is to follow Algorithm 1, however, the network size will change with order, which we avoid using recurrence.

\begin{figure}[htb!]
\centering
\includegraphics[width=0.95\linewidth]{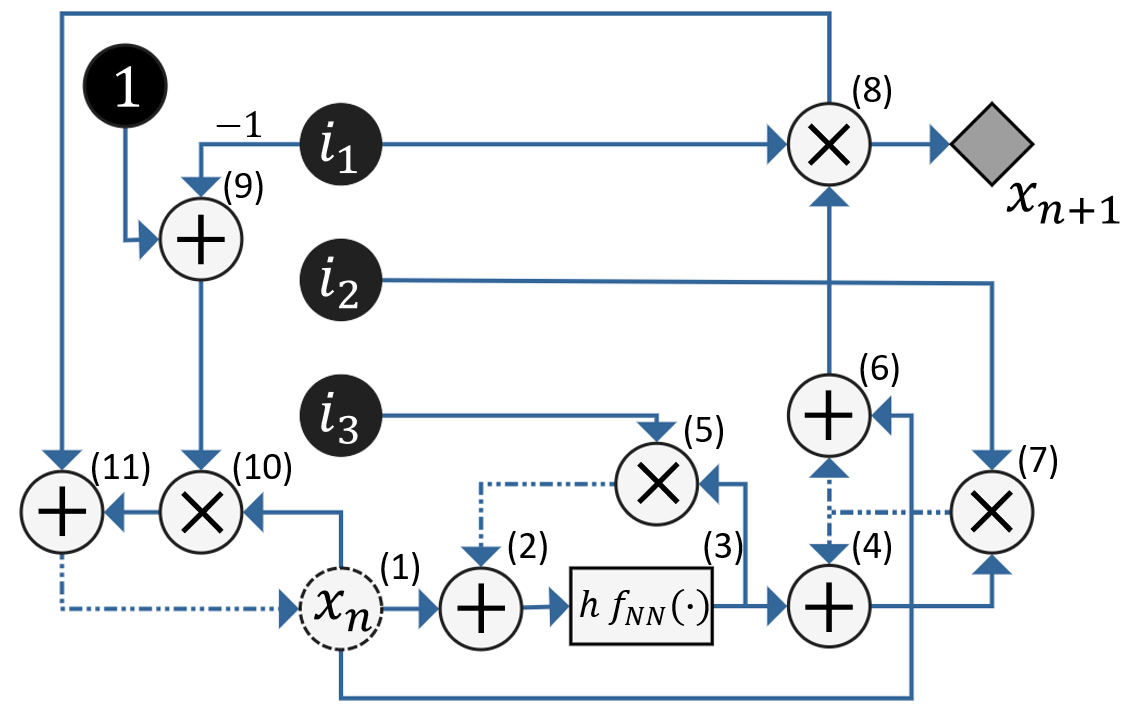}
\caption{\it A neural circuit for an explicit fourth-order Runge Kutta method. The current state $x_n$ enters in the circular node labelled $(1)$ with a dashed border. The numbers to the upper right of each node indicate their order in the computation. This order is not unique, but it is useful to follow for the sake of comprehension. The rectangular node represents the computation of the gradient of a particular dynamical system at the point specified by the input. The black nodes indicate constant coefficients,  $i_1, i_2, i_3$ cycle values every five iterations. This cyclic coefficient input is achieved through a sub-circuit shown in Figure \ref{fig:RK_Coeff}. The dashed edges indicate a delay-- they pass the start node values from the previous iteration to their end nodes, and they hold the new value until the next iteration. The diamond node represents the output. Every five iterations of the circuit produces the next element in the time series. }
\label{fig:RK_Neural_Circuit}
\end{figure}

\begin{figure}[htb!]
\centering
\includegraphics[width=0.65\linewidth]{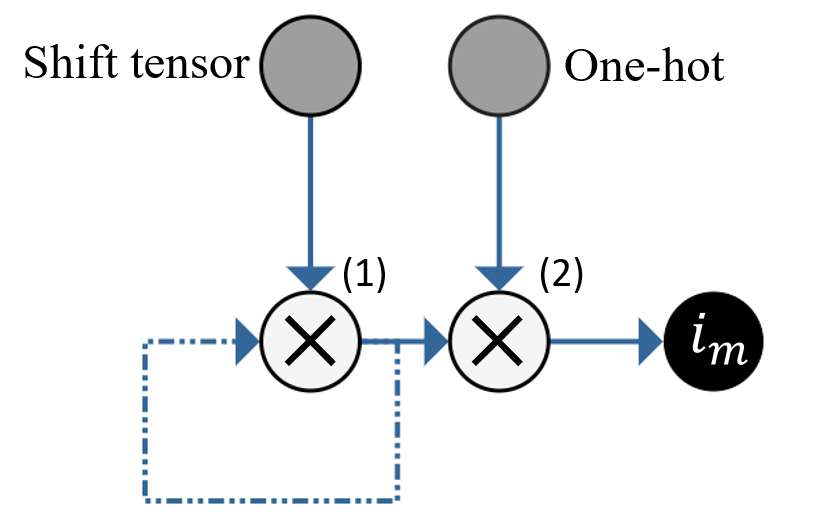}
\caption{\it A neural circuit for the cyclical coefficient input. As an initial state, node $(1)$ contains the initial coefficient array for coefficient $m$. For the Runge-Kutta implementation shown in Figure \ref{fig:RK_Neural_Circuit}, these arrays are $[0, 0, 0, 0, 1]$, $[\frac{1}{2}, 1, 2, \frac{1}{6}, 0]$, and $[\frac{1}{2}, \frac{1}{2}, 1, 0, 0]$ for nodes $i_1$, $i_2$, and $i_3$, respectively, and $m \in \{1,2,3\}.$ A constant one-hot vector multiplies with the coefficient array of node $(1)$ to pass the first element to the black node as the coefficient for that iteration. A shift tensor multiplies with the coefficient array in node $(1)$ to rotate the elements, and the shifted array is passed back for the next iteration. In this manner, the first elements of each coefficient array are cycled.}
\label{fig:RK_Coeff}
\end{figure}

 Consider instead the network depicted in Figure~\ref{fig:RK_Neural_Circuit} and Figure~\ref{fig:RK_Coeff}. In this circuit, coefficients cyclically enter the network through the nodes on the left, and the output for the next time step is computed after five iterations of the system. In this circuit, computation occurs in the order specified by the numbers adjacent to each node. The computations at nodes $(5)$, $(7)$, and $(11)$ are then held by delays in the recurrence until the next iteration. Each node takes the sum or product of its inputs as indicated by the symbols on each node. In the circuit, $f_{NN}$ represents the gradient computation achieved via any neural circuit representing a set of polynomial ODEs. In our example, the Lorenz neural circuit in Figure \ref{fig:Lorenz_Circuit} is implemented for $f_{NN}$, but any neural dynamical system could replace it. Every $5$ iterations, node $(8)$ contains the output that is the next element in the time series. With this construction, given a single initial state, the neural circuit iteratively outputs the time series. The neural circuit in Figure~\ref{fig:RK_Neural_Circuit} is an exact recurrent PolyNet of fixed size that implements explicit fixed-step RK4 shown in Equations~\ref{eq:rk4a}-\ref{eq:rk4b}.

The structure shown in Figure \ref{fig:RK_Neural_Circuit} can be modified slightly to apply to a general Runge-Kutta method. Since we consider general Runge-Methods to integrate ordinary differential equations, the equations take the form

\begin{eqnarray*}
 k_1 &=& h f_{NN}(x_n)\\
 k_2 &=& h f_{NN}(x_n+ a_{21}k_1) \\
 k_3 &=& h f_{NN}(x_n + a_{31}k_1 + a_{32}k_2) \\
 \vdots \\
 k_s  &=& h f_{NN}(x_n + a_{s1}k_1 + a_{s2}k_2 + \hdots + a_{s,s-1}k_{s-1}) \\
 x_{n+1} &=& x_n + \sum_{i=1}^{s}b_i k_i
\end{eqnarray*}
To generalize the structure of our system, we modify node $(5)$ and the associated coefficient array $i_3$ to an array of nodes, one for each $k_i$ where $1\leq i \leq s-1$, such that node $(3)$ will pass the appropriate $k_i$ to each node with a timed one-hot multiplicative weight. Each node will pass the $k_i$ to itself and node $(2)$ with the appropriate multiplicative coefficient such that the input to $(2)$ from array $(5)$ is always $\sum_{j=1}^{i-1}a_{ij}k_j$ at round $i$ for $1\leq i \leq s$. In addition, the coefficient array $i_2$ must change to 
\begin{equation*}
    [\frac{b_1}{b_2}, \frac{b_2}{b_3}, \hdots, \frac{b_{s-1}}{b_s}, b_s, 0].
\end{equation*}
Finally, the coefficient array $i_1$ changes to the length $s+1$ one-hot vector with the single high bit in the final entry. With these changes, the neural architecture for the $4^{\text{th}}$ order Runge-Kutta method applies to general explicit Runge-Kutta methods. Note that the order of approximation can be varied during neural integration merely by increasing or decreasing coefficient sequences without changing the network. This provides a serious advantage in changing stiffness/dynamical regimes.

\subsection{Semi-Implicit Methods}

\begin{figure}[htb!]
\centering
\includegraphics[width=0.95\linewidth]{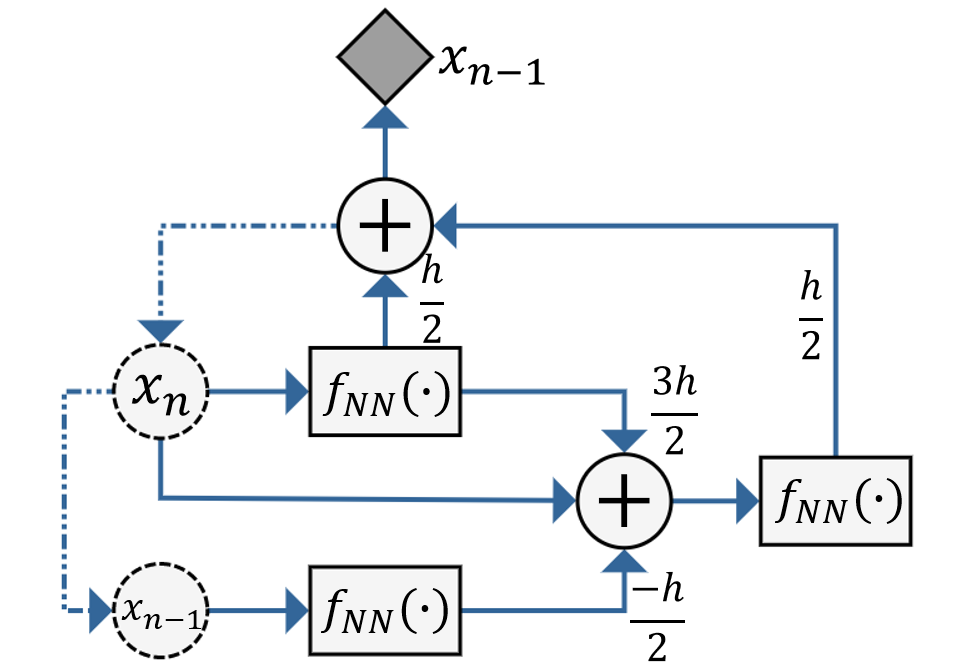}
\caption{\it A neural circuit for a two-step Adams-Bashforth predictor coupled with an Adams-Moulton Corrector.}
\label{fig: Adams-Bashforth}
\end{figure}

To extend the methodology to other numerical integrators, consider the two-step Adams-Bashforth predictor, coupled to a two-step Adams- Moulton Corrector, according to 
\begin{eqnarray*}
 P_{n+1} &=& x_n + \frac{3}{2} h f_{NN}(x_n) -\frac{1}{2}hf_{NN}(x_{n-1})\\
 x_{n+1} &=& x_n + \frac{1}{2}h(f_{NN}(P_{n+1}) +f_{NN}(x_n)).
\end{eqnarray*}
The neural architecture for this numerical integrator is shown in Figure \ref{fig: Adams-Bashforth}, where again any equation $\dot{x} = f_{NN}(x)$ can be integrated by the system.

\begin{figure*}[ht!]
  \begin{subfigure}[t]{0.4\textwidth}
    \caption{}
    \includegraphics[width=\textwidth, valign=t]{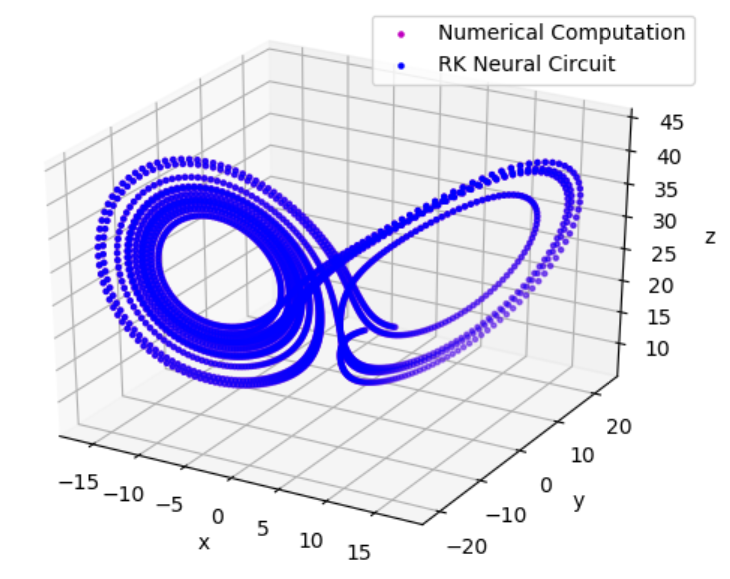}
    \label{fig:Error_a}
  \end{subfigure}
  \begin{subfigure}[t]{0.38\textwidth}
    \caption{}
    \includegraphics[width=\textwidth, valign=t]{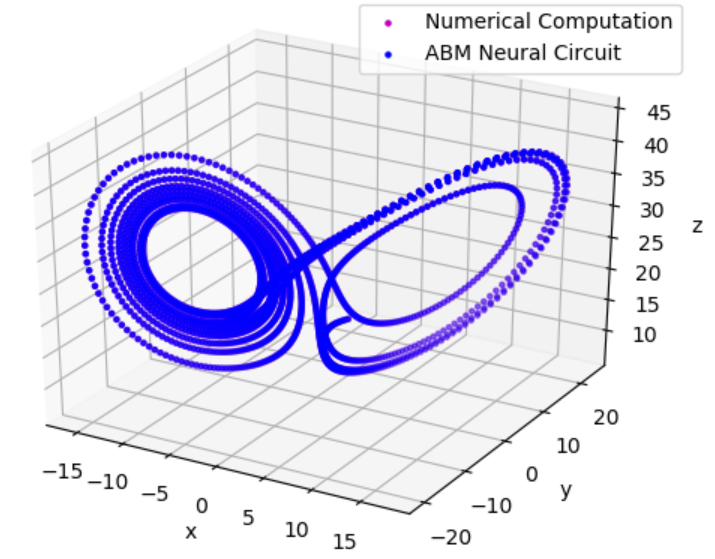}
    \label{fig:Error_AB}
  \end{subfigure}
  \begin{subfigure}[t]{0.4\textwidth}
    \caption{}
    \includegraphics[width=\textwidth]{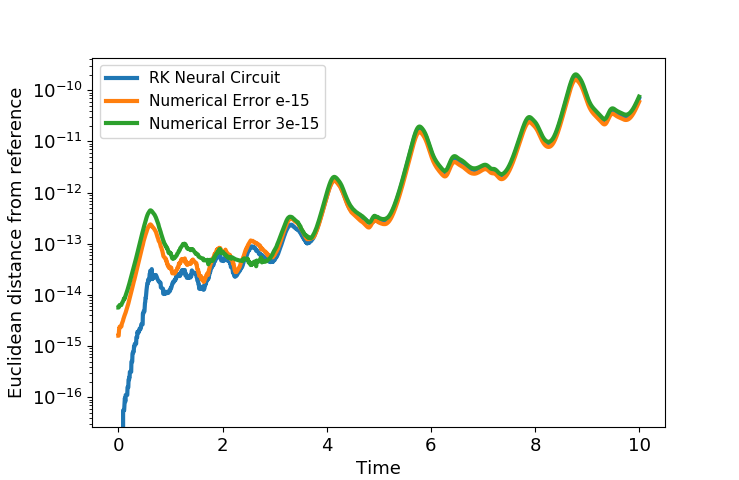}
    \label{fig:Error_b}
  \end{subfigure}
  \begin{subfigure}[t]{0.4\textwidth}
    \caption{}
    \includegraphics[width=\textwidth]{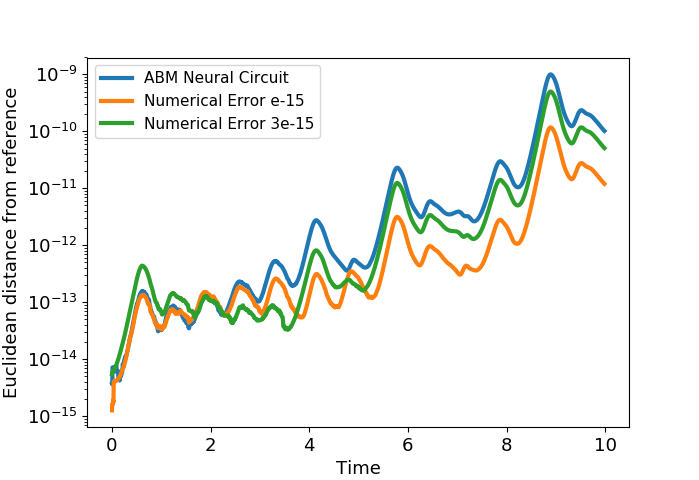}
    \label{fig:Error_c}
  \end{subfigure}
  \caption{\it A comparison of the neural circuit output for both the fourth order Runge-Kutta scheme and the Adams-Bashforth-Moulton corrector to the standard numerical computation for the toy example of the Lorenz system. $\mathbf{(a)}$ A rendering of both the classical numerical computation output for the Lorenz system time series and the Runge-Kutta neural circuit output. $\mathbf{(b)}$ A similar rendering for the Adams-Bashforth-Moulton corrector. $\mathbf{(c)}$ Euclidean distance from the Runge-Kutta neural circuit output to a reference time series generated in the classical RK numerical method. For comparison, the numerical error resulting from initial perturbations of $10^{-15}$ and $3(10^{-15})$ to the classical RK numerical method are shown. $(d)$ Euclidean distance from the Adams-Bashforth-Moulton corrector method output to a reference time series generated in the classical ABM numerical method. For comparison, the numerical error resulting from initial perturbations of $10^{-15}$ and $3\times 10^{-15}$ to the classical ABM numerical method are shown.}
  \label{fig:valid}
\end{figure*}

Figure \ref{fig:valid} depicts a comparison between a fourth-order explicit Runge-Kutta neural integration, a second order Adams-Bashforth-Moulton method neural integration and the corresponding ordinary numerical integration. The computational difference between the neural circuit method and the classical numerical computations were that the neural circuit was implemented with a TensorFlow TensorGraph, and the classical numerical computation was done with a simple function\footnote{All programs were written in Python and executed on a Laptop.}. Unsurprisingly, since the operations are mathematically equivalent, the output time series are exactly equivalent, save computational floating point discrepancies that magnify as a result of the system's chaotic behavior. To demonstrate that the discrepancies are due to floating point computation, we have perturbed the initial condition by $10^{-15}$, which is the precision used for the computation. As a comparison, we have also generated a time series with a perturbation of $3\times 10^{-15}$ to show that there is some level of randomness in the accuracy of the system following an initial perturbation. 

\section{Discussion}
\label{sec:disc}

 Although exact PolyNets are sufficient to establish the correctness of neural integration, two issues remain. The first is computational efficiency.  We posit  that constructing a DtNN allows for faster and more efficient computation than interleaving neural and classical computation. Representing calculations with tensors enables computation on specialized hardware for potentially faster performance. The scaling with  independently executing DtNNs, such as for uncertainty quantification and other applications, can be enormous. Indeed, for nominal ODE integration, such benefit already been shown~\cite{Niemeyer2014}. It remains to be seen whether the benefits apparent for many independent DtNN simulations extend to other problems such as variational Bayes. We hypothesize that the gains could be significant especially if coupled with a low-cost variable-order neural integration, as our circuits allow, is minimal. 

The second issue is of training embedded CtNNs. To discuss this further, consider the network-dynamics table shown in Table~\ref{tab:poly}.  When polynomial  structure is known but coefficients are unknown, {\em PolyNet Identification}\footnote{This term is used in the sense of System Identification in Estimation and Control.} is necessary.  This is accomplished using the discrete adjoint method~\cite{bryson} for DtNN parameter estimation. In the forward pass, the DtNN simulates outputs up to a horizon and, in the backward pass, errors are back-propagated  for  estimation. DtNN learning over a receding horizon relates closely to classical receding horizon model predictive optimal control. Of course, DtNNs can be trained using alternate approaches, particularly using Bayesian filters, and fixed point, fixed interval and fixed lag smoothers~\cite{Ravela2007}. 

\begin{table}[htb!]
    \centering
    \begin{tabular}{|c||c|c|c||}
    \hline 
        \multirow{2}{*}{Network}&\multicolumn{3}{c||}{Dynamics}\\
         &Poly w/ Coeff. & Poly w/o Coeff. & Non-poly  \\
         \hline\hline
        PolyNet & Exact & Estimate & Approx. \\\hline
        Other & Poor & General & N/A \\\hline
    \end{tabular}
    \caption{\it Neural PolyNet and general models in relation to nonlinear dynamics.}
    \label{tab:poly}
\end{table}

  When the dynamics have non-polynomial governing equations, PolyNets are a useful approximation in the sense of Stone-Weierstrass~\cite{weierstrass} theorem enabling approximations of smooth non-linearities.  However, in this case the PolyNet structure (number of monomials, degree, connectivity) is also in error. The challenging problem of {\em PolyNet Structure Learning}\footnote{Used analogously to Structure Learning in Machine Learning and Structural Model Errors in Geophysics.} must be addressed.  However, if parsimony principles are applicable to modeling, then one can gradually increase PolyNet complexity, perhaps by sampling, and apply model selection criteria. This approach is however in stark contrast to structure learning wherein a dense, large network is ``made sparse," a problem both in neural learning as well as graphical models. The systematization of model complexity may enable more tractable structure learning.  
  
  PolyNets are also, we posit, promising theory-guided learning machines~\cite{karpatne19}. Using theory, via governing equations to establish the initial PolyNet makes them ``explainable" and avoids the "black-box" nature typical neural construction. Subsequently, with the arrival of data, PolyNet Identification and PolyNet Structure Learning interleave, following parsimony principles titrating model complexity. In these steps, to be sure, only PolyNet terms are estimated-- the integrator loop remains immutable.  Thus, constructing neural circuits for dynamical systems would potentially allow for inference of the underlying equations of a system while accounting for the known physical behavior. 

  Theory-guided learning, as discussed, offers enormous advantage. Contrast it with the general setting wherein a polynomial is learned directly from data. Using a two-layer neural network similar to the PolyNet, bounds derived by Andoni et al.~\cite{andoni} are show that the problem is substantially worse. This absence of guidance for PolyNet design implies a combinatorial explosion emerging with degree. This in turn causes the number of nodes, required training data, iterations to convergence to all become prohibitive for most applications. 

 When the governing equations are polynomial but PolyNets aren't used, the situation also becomes quite a bit complicated. Neural networks have been argued to be polynomial machines where hidden layers~\cite{PolyReg} control the effective degree.  However, while neural architectures offer enormous modeling flexibility through myriad activation and learning rules, they can be very sub-optimal with respect to number of parameters needed to model a polynomial. For example, a single neuron with {\em tanh} activation poorly models $y=x^2$\footnote{See http://essg.mit.edu/ml/yx2}; sensitivity to noise, weak generalization and no `extrapolation" skill are easily evident. In fact, the Taylor expansion of $tanh$ has no second-degree term. It may be easier to perform "polynomial regression" instead (as~\cite{PolyReg} argue) provided, as discussed in the previous paragraph, the combinatorial explosion of monomials can be managed. We hypothesize that layered architectures alternating compression (by dimensionality reduction)  with nonlinear stretching (by low-degree PolyNets) could be a promising alternative to classical neural circuitry or classical polynomial regression. 
 
 Regardless, given polynomial equations and the insistence on using a standard neural network architecture (e.g. feed-forward with smooth nonlinear activation), we show elsewhere that bounds on neural structure can be established~\cite{ravela19}. These bounds have been experimentally validated~\cite{ziwei}, but naturally are higher than PolyNets provide. For example, for the Lorenz system, exactly $2$ PolyNet nodes are needed, however, a feed-forward network with $tanh$ activation, $6$ nodes appear to be necessary. Andoni et al.'s bounds are prohibitively weaker.
 
\section{Conclusions}
In this paper we demonstrate neural integration: solving neural ordinary differential equations as neural computation. Neural integration is carried out by Discrete-time Neural Networks (DtNN). For a provided Neural-ODE, we construct DtNNs for explicit fixed-step Runge-Kutta methods of any order. The DtNNs are recurrent PolyNets and remain of fixed size irrespective of order. We also construct DtNN for an second-order Adams-Bashforth predictor coupled to a second-order Adams-Moulton corrector. In both these cases, we show that neural integration is correct for any correct embedded Continuous-time Neural Network (CtNN). We further demonstrate this numerically using exact PolyNet for the Lorenz-63 attractor, and showing convergence between neural and numerical integration up to numerical representation. We argue that PolyNets offer a  for theory-guided modeling because the availability.

Our focus in this first of a three-part paper is to simply demonstrate neural integration.  The next paper focuses PolyNet identification and the third paper focuses on PolyNet adaptation. We also seek to apply this methodology to Systems Dynamics and Optimization (SDO) problems in the environment and for the design of new adaptive autopilots for environmental monitoring.

\label{sec:concl}

\section*{Acknowledgments} Ms. Trautner is an undergraduate researcher at ESSG advised by Dr. Ravela. Support from ONR grant N00014-19-1-2273, the MIT Environmental Solutions Initiative, the Maryanne and John Montrym Fund, and the MIT Lincoln Laboratory are gratefully acknowledged. Discussions with members of the Earth Signals and Systems group is also gratefully acknowledged. 
\appendix

\bibliographystyle{aipauth4-1}
\bibliography{secondary_general_project}

\end{document}